\documentclass{article}

\usepackage{microtype}
\usepackage{graphicx}
\usepackage{subcaption}
\usepackage{booktabs} 
\usepackage{amsmath}
\usepackage[mathscr]{euscript}
\usepackage{amssymb}
\usepackage{algorithm}
\usepackage{enumitem}
\usepackage{multirow}

\def\sQ{{\mathbb{Q}}}
\newcommand{\IFTHEN}[2]{
	\STATE \algorithmicif\ #1\ \algorithmicthen\ #2}
\newcommand{\IFTHENELSE}[3]{
	\STATE \algorithmicif\ #1\ \algorithmicthen\ #2\ \algorithmicelse\ #3}

\usepackage{hyperref}

\usepackage[accepted]{icml2019_arxiv}

\icmltitlerunning{Hierarchically Clustered Representation Learning}

\graphicspath{{fig/}}

\begin{document}

\twocolumn[
\icmltitle{Hierarchically Clustered Representation Learning}



\icmlsetsymbol{equal}{*}

\begin{icmlauthorlist}
\icmlauthor{Su-Jin Shin}{kaist}
\icmlauthor{Kyungwoo Song}{kaist}
\icmlauthor{Il-Chul Moon}{kaist}
\end{icmlauthorlist}

\icmlaffiliation{kaist}{Department of Industrial and Systems Engineering, KAIST, Daejeon, Republic of Korea}

\icmlcorrespondingauthor{Il-Chul Moon}{icmoon@kaist.ac.kr}

\icmlkeywords{Machine Learning, ICML}

\vskip 0.3in
]



\printAffiliationsAndNotice{}  

\begin{abstract}
The joint optimization of representation learning and clustering in the embedding space has experienced a breakthrough in recent years. In spite of the advance, clustering with representation learning has been limited to flat-level categories, which often involves cohesive clustering with a focus on instance relations. To overcome the limitations of flat clustering, we introduce \textit{hierarchically-clustered} representation learning (HCRL), which simultaneously optimizes representation learning and hierarchical clustering in the embedding space. Compared with a few prior works, HCRL firstly attempts to consider a generation of deep embeddings from every component of the hierarchy, not just leaf components. In addition to obtaining hierarchically clustered embeddings, we can reconstruct data by the various abstraction levels, infer the intrinsic hierarchical structure, and learn the level-proportion features. We conducted evaluations with image and text domains, and our quantitative analyses showed competent likelihoods and the best accuracies compared with the baselines.
\end{abstract}

\section{Introduction}

%

\begin{figure}[t]
\centerline{\includegraphics[width=1.1\columnwidth]{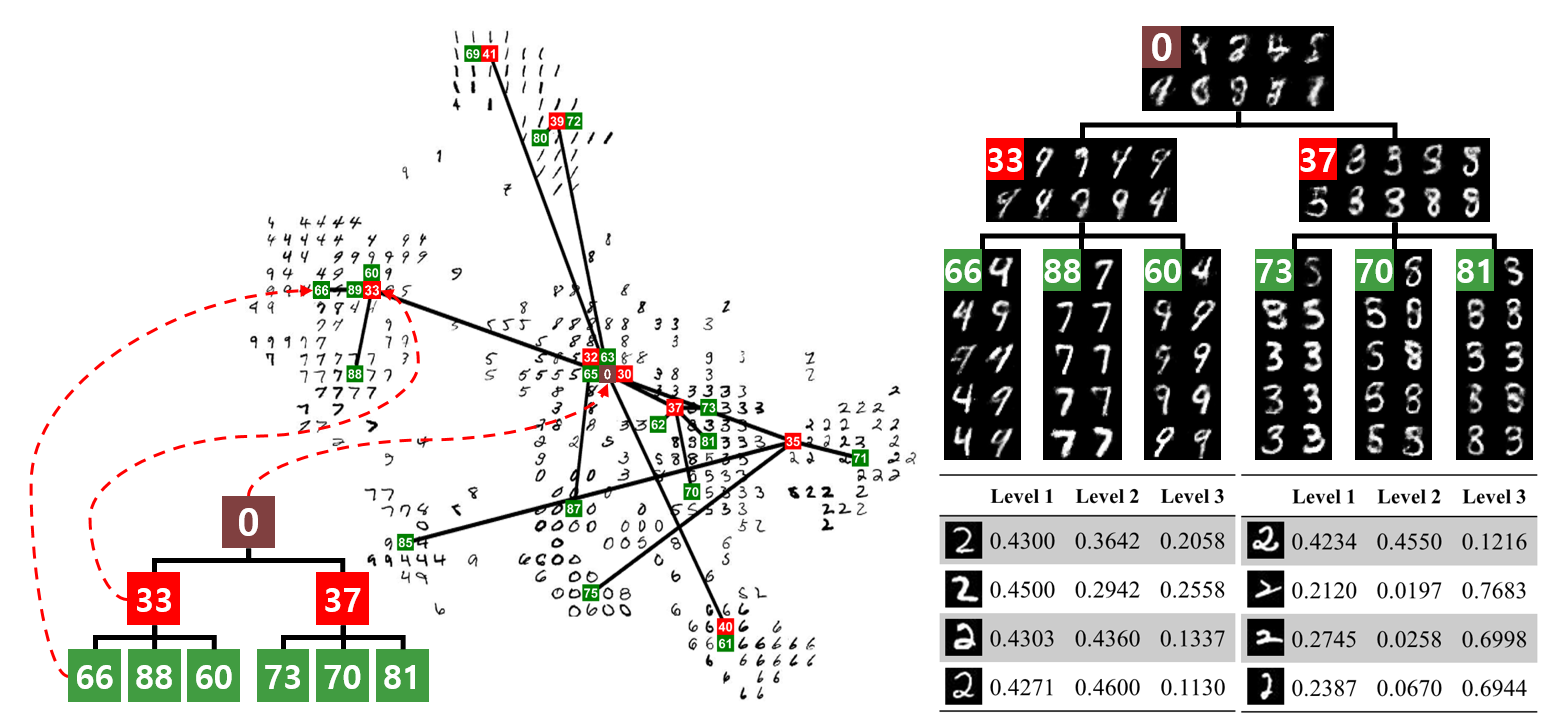}}
\caption{Example of hierarchically clustered embeddings on MNIST with three levels of hierarchy (left), the generated digits from the hierarchical Gaussian mixture components (top right), and the extracted level proportion features (bottom right). We marked the mean of a Gaussian mixture component with the colored square, and the digit written inside the square refers to the unique index of the mixture component.}
\label{fig:mnist}
\end{figure}

\begin{figure*}[htbp]
\centerline{\includegraphics[width=1.0\textwidth]{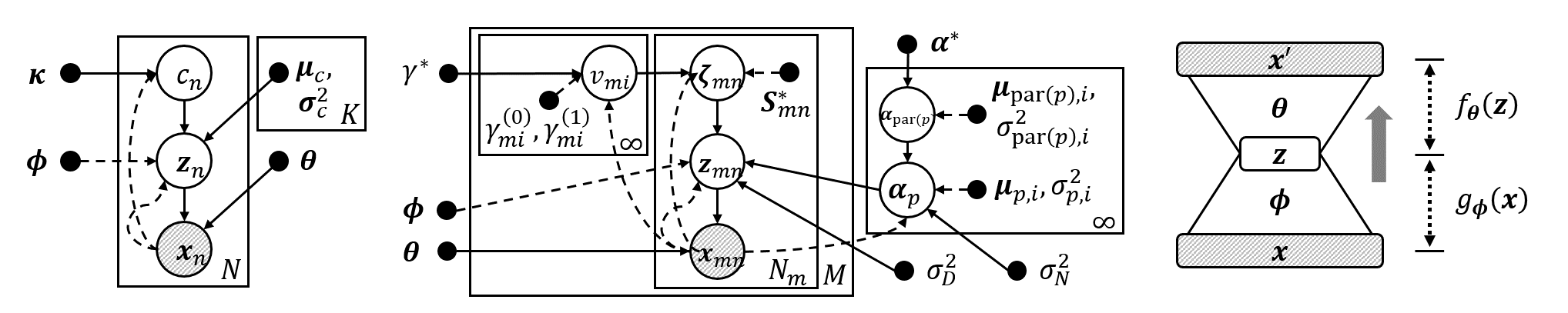}}
     \caption{Graphical representation of VaDE \citep{jiang2016variational} (left), VAE-nCRP \citep{goyal2017nonparametric} (center), and neural architecture of both models (right). In the graphical representation, the white/shaded circles represent latent/observed variables. The black dots indicate hyper or variational parameters. The solid lines represent a generative model, and dashed lines represent a variational approximation. A rectangle box means a repetition for the number of times denoted by the bottom right of the box.}
    \label{fig:baseline}
\end{figure*}

\textit{Clustering} is one of the most traditional and frequently used machine learning tasks. Clustering models are designed to represent intrinsic data structures, such as latent Dirichlet allocation \citep{blei2003latent}. 
The recent development of \textit{representation learning} has contributed to generalizing model feature engineering, which also enhances data representation \citep{bengio2013representation}. Therefore, representation learning has been merged into the clustering models, e.g., variational deep embedding (VaDE) \citep{jiang2016variational}. 

Autoencoder \citep{rumelhart1985learning} is a typical neural network for unsupervised representation learning and achieves a non-linear mapping from a input space to a embedding space by minimizing reconstruction errors. To turn the embeddings into random variables, a variational autoencoder (VAE) \citep{kingma2013auto} places a Gaussian prior on the embeddings. The autoencoder, whether it is probabilistic or not, has a limitation in reflecting the intrinsic hierarchical structure of data. For instance, VAE assuming a single Gaussian prior needs to be expanded to suggest an elaborate clustering structure.

Due to the limitations of modeling the cluster structure with autoencoders, prior works combine the autoencoder and the clustering algorithm. While some early cases pipeline just two models, e.g., \citet{huang2014deep}, a typical merging approach is to model an additional loss, such as a clustering loss, in the autoencoders \citep{xie2016unsupervised,guo2017improved,yang2017towards,nalisnick2016approximate,chu2017stacked,jiang2016variational}. These suggestions exhibit gains from unifying the encoding and the clustering, yet they remain at the parametric and flat-structured clustering. A more recent development releases the previous constraints by using the nonparametric Bayesian approach. For example, the infinite mixture of VAEs (IMVAE) \citep{abbasnejad2017infinite} explores the infinite space for VAE mixtures by looking for an adequate embedding space through sampling, such as the Chinese restaurant process (CRP). Whereas IMVAE remains at the flat-structured clustering, VAE-nested CRP (VAE-nCRP) \citep{goyal2017nonparametric} captures a more complex structure, i.e., a hierarchical structure of the data, by adopting the nested Chinese restaurant process (nCRP) prior \citep{griffiths2004hierarchical} into the cluster assignment of the Gaussian mixture model.

Hierarchical mixture density estimation \citep{vasconcelos1999learning}, where all internal and leaf components are directly modeled to generate data, is a flexible framework for hierarchical mixture modeling, such as hierarchical topic modeling \citep{mimno2007mixtures,griffiths2004hierarchical}, with regard to the learning of the internal components. This paper proposes hierarchically clustered representation learning (HCRL) that is a joint model of 1) nonparametric Bayesian hierarchical clustering, and 2) representation learning with neural networks. HCRL extends a previous work on merging flat clustering and representation learning, i.e., VaDE, by incorporating inter-cluster relation modelings. 

Specifically, HCRL jointly optimizes soft-divisive hierarchical clustering in an embedding space from VAE via two mechanisms. First, HCRL includes a hierarchical-versioned Gaussian mixture model (HGMM) with a mixture of hierarchically organized Gaussian distributions. Then, HCRL sets the prior of embeddings by adopting the generative processes of HGMM. Second, to handle a dynamic hierarchy structure dealing with the clusters of unequal sizes, we explore the infinite hierarchy space by exploiting an nCRP prior. 
These mechanisms are fused as a unified objective function; this is done rather than concatenating the two distinct models of clustering and autoencoding.

We developed two variations of HCRL, called HCRL1 and HCRL2, where HCRL2 extends HCRL1 by the flexible modeling on the level proportion. The quantitative evaluations focus on density estimation quality and hierarchical clustering accuracy, which shows that HCRL2 have competent likelihoods and the best accuracies compared with the baselines. When we observe our results qualitatively, we visualize 1) the hierarchical clusterings, 2) the embeddings under the hierarchy modeling, and 3) the generated images from each Gaussian mixture component, as shown in Figure \ref{fig:mnist}. These experiments were conducted by crossing the data domains of texts and images, so our benchmark datasets include MNIST, CIFAR-100, RCV1\_v2, and 20Newsgroups.

\section{Preliminaries}

\subsection{Variational Deep Embedding}

Figure \ref{fig:baseline} presents a graphical representation and a neural architecture of VaDE \citep{jiang2016variational}. The model parameters of $\boldsymbol{\kappa}$, $\boldsymbol{\mu}_{1:K}$, and $\boldsymbol{\sigma}^2_{1:K}$, which are a proportion, means, and covariances of mixture components, respectively, are declared outside of the neural network.
VaDE trains model parameters to maximize the lower bound of marginal log likelihoods via the mean-field variational inference \citep{jordan1999introduction}.
VaDE uses the Gaussian mixture model (GMM) as the prior, whereas VAE assumes a single standard Gaussian distribution on embeddings. Following the generative process of GMM, VaDE assumes that 1) the embedding draws a cluster assignment, and 2) the embedding is generated from the selected Gaussian mixture component.

VaDE uses an amortized inference as VAE, with a generative and inference networks; $\mathscr{L}(\boldsymbol{x})$ in Equation \ref{eq:vade} denotes the evidence lower bound (ELBO), which is the lower bound on the log likelihood. It should be noted that VaDE merges the ELBO of VAE with the likelihood of GMM.
\begin{multline}
\log p(\boldsymbol{x}) \geq \mathscr{L}(\boldsymbol{x}) 
= \mathbb{E}_{q}\biggl[\log \frac{p(\boldsymbol{c,z,x})}{q(\boldsymbol{c,z|x})}\biggr] \\
= \mathbb{E}_{q}\biggl[\log \prod_{c=1}^K \frac{\kappa_{c} \mathcal{N}(\boldsymbol{z}|\boldsymbol{\mu}_{c},\boldsymbol{\sigma}^{2}_{c}\boldsymbol{I}_{J})}{p(c|\boldsymbol{z}) \mathcal{N}(\boldsymbol{z}|\widetilde{\boldsymbol{\mu}},\widetilde{\boldsymbol{\sigma}}^2\boldsymbol{I}_{J})} + \log p(\boldsymbol{x|z}) \biggr]
\label{eq:vade}
\end{multline}

\subsection{Variational Autoencoder nested Chinese Restaurant Process}

VAE-nCRP uses the nonparametric Bayesian prior for learning tree-based hierarchies, the nCRP \citep{griffiths2004hierarchical}, so the representation could be hierarchically organized. The nCRP prior defines the distributions over children components for each parent component, recursively in a top-down way. The variational inference of the nCRP can be formalized by the nested stick-breaking construction \citep{wang2009variational}, which is also kept in the VAE setting.
The distribution over paths on the hierarchy is defined as being proportional to the product of weights corresponding to the nodes lying in each path. The weight, $\pi_i$, for the $i$-th node follows the Griffiths-Engen-McCloskey (GEM) distribution \citep{pitman2002combinatorial}, where $\pi_i$ is constructed as $\pi_i=v_i \prod_{j=1}^{i-1} (1-v_j), v_i \sim \operatorname{Beta}(1,\gamma)$ by a stick-breaking process.  Since the nCRP provides the ELBO with the nested stick-breaking process, VAE-nCRP has a unified ELBO of VAE and the nCRP in Equation \ref{eq:vaencrp}.

\begin{multline}
\\[-4em]
\mathscr{L}(\boldsymbol{x})
=\mathbb{E}_{q}\biggl[ \log \frac{p(\boldsymbol{v})}{q(\boldsymbol{v}|\boldsymbol{x})} 
+ \log p(\boldsymbol{x}|\boldsymbol{z}) +  \log \Big\{ \frac{p(\boldsymbol{\zeta}|\boldsymbol{v})}{q(\boldsymbol{\zeta}|\boldsymbol{x})} \\
 \underbrace{\frac{p(\boldsymbol{\alpha}_{\textup{par}(p)}|\boldsymbol{\alpha}^{*})p(\boldsymbol{\alpha}_p|\boldsymbol{\alpha}_{\textup{par}(p)},\sigma_N^2)}{q(\boldsymbol{\alpha}_p,\boldsymbol{\alpha}_{\textup{par}(p)}|\boldsymbol{x})}}_{(3.1)} \underbrace{\frac{p(\boldsymbol{z}|\boldsymbol{\alpha}_{p},{\boldsymbol{\zeta}},\sigma_D^2)}{q(\boldsymbol{z}|\boldsymbol{x})}}_{(3.2)} \Big\} \biggr]
\label{eq:vaencrp}
\end{multline}
Given the ELBO of VAE-nCRP, we recognized the potential improvements. First, term (3.1) is for modeling the hierarchical relationship among clusters, i.e., each child is generated from its parent. VAE-nCRP trade-off is the direct dependency modeling among clusters against the mean-field approximation. This modeling may reveal that the higher clusters in the hierarchy are more difficult to train. Second, in term (3.2), leaf mixture components generate embeddings, which implies that only leaf clusters have direct summarization ability for sub-populations. Additionally, in term (3.2), variance parameter $\sigma_D^2$ is modeled as the hyperparameter shared by all clusters. In other words, only with $J$-dimensional parameters, $\boldsymbol{\alpha}$, for the leaf mixture components, the local density modeling without variance parameters has a critical disadvantage.

For all of these weaknesses, we were able to compensate with the level proportion modeling and HGMM prior. The level assignment generated from the level proportion allows a data instance to select among all mixture components. We do not need direct dependency modeling between the parents and their children because all internal mixture components also generate embeddings. 

%

\section{Proposed Models}



\subsection{Generative Process}
\label{sec:generative}

\begin{figure*}[htbp]
\centerline{\includegraphics[width=1.0\textwidth]{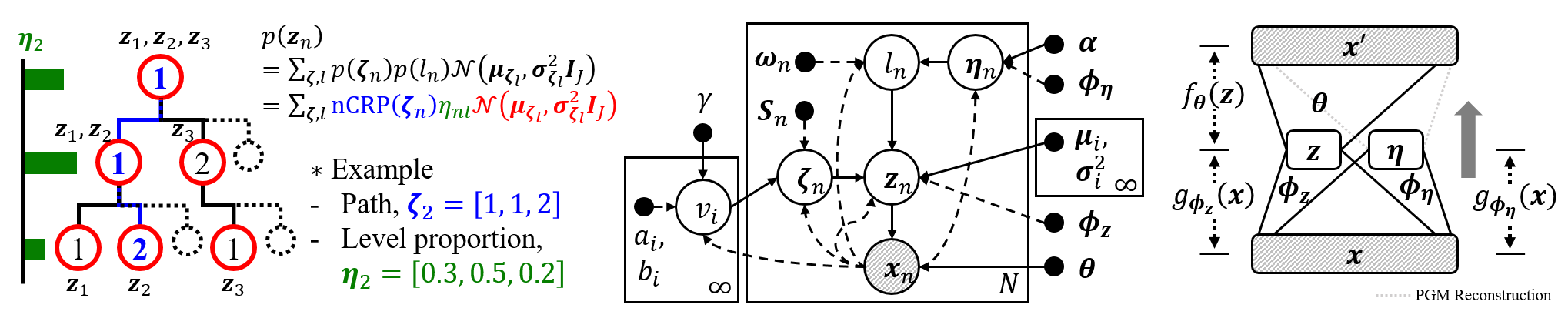}}
\caption{A simple depiction (left) of the key notations, where each numbered circle refers to the corresponding Gaussian mixture component. The graphical representation (center) and the neural architecture (right) of our proposed model, HCRL2. The neural architecture of HCRL2 consists of two probabilistic encoder networks, $g_{\boldsymbol{\phi}_{\boldsymbol{\eta}}}$ and $g_{\boldsymbol{\phi}_{\boldsymbol{z}}}$, and one probabilistic decoder network, $f_{\boldsymbol{\theta}}$.}
\label{fig:hcrl2}
\end{figure*}

We developed two models for the hierarchically clustered representation learning; HCRL1 and HCRL2. The generative processes of the presented models resemble the generative process of hierarchical clusterings, such as the hierarchical latent Dirichlet allocation \citep{griffiths2004hierarchical}. In detail, the generative process departs from selecting a path $\boldsymbol{\zeta}$, from the nCRP prior (Phase 1). Then, we sample a level proportion (Phase 2) and a level, $l$ (Phase 3), from the sampled level proportion to find the mixture component in the path, and this component of $\zeta_l$ provides the Gaussian distribution for the latent representation (Phase 4). Finally, the latent representation is exploited to generate an observed datapoint (Phase 5). The first subfigure of Figure \ref{fig:hcrl2} depicts the generative process with the specific notations.

The level proportion of Phase 2 is commonly modeled as the group-specific variable in the topic modeling. To adapt the level proportion for our non-grouped setting, we considered two modeling assumptions on the level proportion: 1) globally defined the level proportion which is shared by all data instances, which characterizes HCRL1, and 2) locally defined, i.e., data-specific level proportion, which is a distinction of HCRL2 from HCRL1. Similar to the latter assumption, several recently proposed models also define a data-specific mixture membership over the mixture components \citep{zhang2018automatic,ji2016spatially}.

\begin{figure}[htbp]
\centerline{\includegraphics[width=0.75\columnwidth]{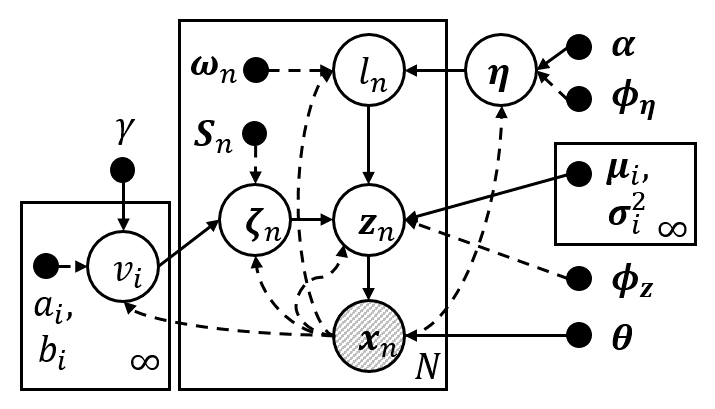}} 
\caption{Graphical representation of HCRL1}
\label{fig:hcrl1}
\end{figure}

The below formulas are the generative process of HCRL2 with its density functions, where the level proportion is generated by a data instance. In addition, Figure \ref{fig:hcrl1} and \ref{fig:hcrl2} illustrate graphical representations of HCRL1 and HCRL2, respectively, and the graphical representations are corresponding to the described generative process. The generative process also presents our formalization of our prior distributions, denoted as $p(\cdot)$, and variational distributions, denoted as $q(\cdot)$, by generation phases. The variational distributions are used for the mean-field variational inference \citep{jordan1999introduction} as detailed in Section \ref{sec:varinf}.

\newcommand{\noindentitem}{\setlength\itemindent{-2pt}}
\begin{enumerate}[noitemsep,topsep=0pt]
	\item Choose a path $\boldsymbol{\zeta} \sim \operatorname{nCRP}(\boldsymbol{\zeta}|\gamma)$
	\begin{itemize}[noitemsep,topsep=0pt]
	\item $p(\boldsymbol{\zeta})=\prod_{l=1}^{L}\pi_{1,\zeta_{2},...,\zeta_{l}}$ 
	where $\pi_{1,\zeta_2,...,\zeta_l} = \prod_{l'=1}^{l}\{v_{1,\zeta_2,...,\zeta_{l'}}(\prod_{j=1}^{\zeta_{l'}-1}(1-v_{1,\zeta_2,...,j}))\} $
	 \item $q(\boldsymbol{\zeta}|\boldsymbol{x})
	\propto S_{\overline{\boldsymbol{\zeta}}} \triangleq \sum_{\boldsymbol{\zeta} \in \textup{child}(\overline{\boldsymbol{\zeta}})}S_{\boldsymbol{\zeta}}  $
	\end{itemize}
	\item Choose a level proportion $\boldsymbol{\eta} \sim \operatorname{Dirichlet}(\boldsymbol{\eta}|\boldsymbol{\alpha})$
	\begin{itemize}[noitemsep,topsep=0pt]
		\item $ p(\boldsymbol{\eta})=\operatorname{Dirichlet}(\boldsymbol{\eta}|\boldsymbol{\alpha}) $
		\item $ q_{\boldsymbol{\phi}_{\boldsymbol{\eta}}}(\boldsymbol{\eta}|\boldsymbol{x})=\operatorname{Dirichlet}(\boldsymbol{\eta}|\widetilde{\boldsymbol{\alpha}}) \\ \approx \operatorname{LogisticNormal}(\boldsymbol{\eta}|\widetilde{\boldsymbol{\mu}}_{\boldsymbol{\eta}},\widetilde{\boldsymbol{\sigma}}^{2}_{\boldsymbol{\eta}} \boldsymbol{I}_L) $ \\
		where $[\widetilde{\boldsymbol{\mu}}_{\boldsymbol{\eta}} ; \log \widetilde{\boldsymbol{\sigma}}^{2}_{\boldsymbol{\eta}}]=g_{\boldsymbol{\phi}_{\boldsymbol{\eta}}}(\boldsymbol{x})$, \\ $ \widetilde{\alpha}_{l} = \frac{1}{\widetilde{\sigma}^2_{\eta_{l}}}(1-\frac{2}{L}+\frac{e^{-\widetilde{\mu}_{\eta_l}}}{L^2}\sum_{l^{'}}e^{{-\widetilde{\mu}_{\eta_{l^{'}}}}}) $
	\end{itemize}	
	\item Choose a level $l \sim \operatorname{Multinomial}(l|\boldsymbol{\eta}) $
	\begin{itemize}[noitemsep,topsep=0pt]
		\item $p(l)=\operatorname{Multinomial}(\boldsymbol{\eta})$
		\item $q(l|\boldsymbol{x}) = \operatorname{Multinomial}(l|\boldsymbol{\omega}) $
		\noindentitem{\item[] where $\omega_{l} \propto \exp \Big\{ \sum_{\boldsymbol{\zeta}}S_{\boldsymbol{\zeta}} \Big( \sum_{j=1}^{J}-\frac{1}{2}\log (2\pi\sigma^2_{\zeta_{l},j}) $} 
		\noindentitem{\item[] $-\frac{\widetilde{\sigma}^2_{z_j}}{2 \sigma^2_{{\zeta}_{l},j}} 
			- \frac{(\widetilde{\mu}_{z_{j}}-\mu_{{\zeta}_{l},j})^2}{2 \sigma^2_{{\zeta}_{l},j}} \Big) + \psi(\widetilde{\alpha}_l)-\psi(\widetilde{\alpha}_0)\Big\} $}
	\end{itemize}	
	\item Choose a latent representation $\boldsymbol{z} \sim \mathcal{N}(\boldsymbol{z}|\boldsymbol{\mu}_{\zeta_{l}},\boldsymbol{\sigma}^{2}_{\zeta_{l}}\boldsymbol{I}_J) $
	\begin{itemize}[noitemsep,topsep=0pt]
		\item $ p(\boldsymbol{z}) = \sum_{\boldsymbol{\zeta},l}p(\boldsymbol{\zeta}|{\gamma}) \cdot \eta_{l} \cdot \mathcal{N}(\boldsymbol{z}|\boldsymbol{\mu}_{\zeta_{l}},\boldsymbol{\sigma}^{2}_{\zeta_{l}}\boldsymbol{I}_J) $
		\item $ q_{\boldsymbol{\phi}_{\boldsymbol{z}}}(\boldsymbol{z}|\boldsymbol{x}) = \mathcal{N}(\boldsymbol{z}|\widetilde{\boldsymbol{\mu}}_{\boldsymbol{z}},\widetilde{\boldsymbol{\sigma}}^2_{\boldsymbol{z}}\boldsymbol{I}_J) $ \\
		where $[\widetilde{\boldsymbol{\mu}}_{\boldsymbol{z}};\log\widetilde{\boldsymbol{\sigma}}^2_{\boldsymbol{z}}] = g_{\boldsymbol{\phi}_{\boldsymbol{z}}}(\boldsymbol{x})$
	\end{itemize}
	\item Choose an observed datapoint $\boldsymbol{x} \sim \mathcal{N}\left(\boldsymbol{x} | \boldsymbol{\mu}_{\boldsymbol{x}},\boldsymbol{\sigma}_{\boldsymbol{x}}^2\boldsymbol{I}_D\right)$ where $[\boldsymbol{\mu}_{\boldsymbol{x}};\log \boldsymbol{\sigma}_{\boldsymbol{x}}^2] = f_{\boldsymbol{\theta}}({\boldsymbol{z}})$\footnote{We introduce the sample distribution for the real-valued data instances, and supplementary material Section 6 provides the binary case as well, which we use for MNIST.}
\end{enumerate}



\subsection{Neural Architecture}

The discrepancy in prior assumptions on the level assignment leads to the different neural architectures. The neural architecture of HCRL1 is a standard variational autoencoder, while the neural architecture of HCRL2 consists of two probabilistic encoders on $\boldsymbol{z}$ and $\boldsymbol{\eta}$, and one probabilistic decoder on $\boldsymbol{z}$ as shown in the right part of Figure \ref{fig:hcrl2}. We designed the probabilistic encoder on $\boldsymbol{\eta}$ for inferring the variational posterior of data-specific level proportion. The unbalanced architecture originates from our modeling assumption of $p(\boldsymbol{x}|\boldsymbol{z})$, not $p(\boldsymbol{x}|\boldsymbol{z}, \boldsymbol{\eta})$. 

One may be puzzled by the lack of the generative network of $\boldsymbol{\eta}$, but $\boldsymbol{\eta}$ is used for the hierarchy construction in the nCRP that is a part of the previous section. In detail, $\boldsymbol{\eta}$ is a random variable of the level proportion in Phase 2 of the generative process. The sampling of $\boldsymbol{\eta}$ and $\boldsymbol{\zeta}$ reflects in the selecting a Gaussian mixture component in Phase 4, and the latent vector $\boldsymbol{z}$ becomes an indicator of a data instance, $\boldsymbol{x}$. Therefore, the sampling of $\boldsymbol{\eta}$ from the neural network is linked to the probabilistic modeling of $\boldsymbol{x}$, so the probabilistic model substitutes for creating a generative network from $\boldsymbol{\eta}$ to $\boldsymbol{x}$. 

Considering $\boldsymbol{\eta}$ in HCRL, the inference network is given, but the generative network was replaced by the generative process of the graphical model. If we imagine a balanced structure, then the generative process needs to be fully described by the neural network, but the complex interaction within the hierarchy makes a complex neural network structure. Therefore, the neural network structure in Figure \ref{fig:hcrl2} may disguise that the structure misses the reconstruction learning on $\boldsymbol{\eta}$, but the reconstruction has been reflected in the PGM side of learning. This is also a difference between (VaDE, VAE-nCRP) and HCRL because VaDE and VAE-nCRP adhere to the balanced autoencoder structure.  
We call this reconstruction process, which is inherently a generative process of the traditional probabilistic graphical model (PGM), \textit{PGM reconstruction} (see the decoding neural network part of Figure \ref{fig:hcrl2}). 

\subsection{Mean-Field Variational Inference}
\label{sec:varinf}
The formal specification can be a factorized probabilistic model as Equation \ref{eq:hcrlgen}, which is based on HCRL2. In the case of HCRL1, $\boldsymbol{\eta}_n$ should be changed to $\boldsymbol{\eta}$ and be placed outside the product over $n$.

\begin{multline}
\\[-4em]
p(\mathbf{\Phi},\boldsymbol{x}) 
=\prod_{j \notin \mathscr{M}_{T}}p(v_j|\gamma) \times \prod_{i \in \mathscr{M}_{T}}p(v_i|\gamma)  \times \	\qquad\quad \\
 \prod_{n=1}^{N}p(\boldsymbol{\zeta}_n|\boldsymbol{v})p(\boldsymbol{\eta}_n|\boldsymbol{\alpha})p(l_n|\boldsymbol{\eta}_n)p(\boldsymbol{z}_n|\boldsymbol{\zeta}_n,l_n)p_{\boldsymbol{\theta}}(\boldsymbol{x}_n|\boldsymbol{z}_n)
\label{eq:hcrlgen}
\raisetag{12pt} 
\end{multline}
where $\mathbf{\Phi} =\{ \boldsymbol{v},\boldsymbol{\zeta},\boldsymbol{\eta},\boldsymbol{l},\boldsymbol{z} \}$ denotes the set of latent variables, and $\mathscr{M}_{T}$ denotes the set of all nodes in tree $T$. The proportion and assignment on the mixture components for the $n$-th data instance are modeled by $\boldsymbol{\zeta}_n$ as a path assignment; $\boldsymbol{\eta}_n$ as a level proportion; and $l_n$ as a level assignment. $v$ is a Beta draw used in the stick-breaking construction. We assume that the variational distributions of HCRL2 are as Equation \ref{eq:hcrlinf} by the mean-field approximation. In HCRL1, we also assume the mean-field variational distributions, and therefore, $\boldsymbol{\eta}_n$ should be replaced by $\boldsymbol{\eta}$ and be outside the product over $n$.
\begin{multline}
q(\mathbf{\Phi}|\boldsymbol{x})
= \prod_{j \notin \mathscr{M}_{T}}p(v_j|\gamma) \times \prod_{i \in \mathscr{M}_{T}} q(v_i|a_i,b_i) \times \\
\prod_{n=1}^{N} q(\boldsymbol{\zeta}_n|\boldsymbol{x}_{n}) q_{\boldsymbol{\phi}_{\boldsymbol{\eta}}}(\boldsymbol{\eta}_n|\boldsymbol{x}_{n}) q(l_n|\boldsymbol{\omega}_n,\boldsymbol{x}_n) q_{\boldsymbol{\phi}_{\boldsymbol{z}}}(\boldsymbol{z}_n|\boldsymbol{x}_n) 
\label{eq:hcrlinf}
\end{multline}
where $q_{\boldsymbol{\phi}_{\boldsymbol{\eta}}}(\boldsymbol{\eta}_n|\boldsymbol{x}_{n})$ and $q_{\boldsymbol{\phi}_{\boldsymbol{z}}}(\boldsymbol{z}_n|\boldsymbol{x}_n)$ should be noted because these two variational distributions follow the amortized inference of VAE. $q(\boldsymbol{\zeta}|\boldsymbol{x}) \propto S_{\overline{\boldsymbol{\zeta}}} \triangleq \sum_{\boldsymbol{\zeta} \in \textup{child}(\overline{\boldsymbol{\zeta}})}S_{\boldsymbol{\zeta}}$ is the variational distribution over path $\boldsymbol{\zeta}$, where $\operatorname{child}(\overline{\boldsymbol{\zeta}})$ means the set of all full paths that are not in $T$ but include $\overline{\boldsymbol{\zeta}}$ as a sub path. 
Because we specified both generative and variational distributions, we define the ELBO of HCRL2, $\mathscr{L}=\mathbb{E}_{q}\left [ \log \frac{p(\mathbf{\Phi},\boldsymbol{x})}{q(\mathbf{\Phi}|\boldsymbol{x})} \right ]$, in Equation \ref{eq:hcrl}. Supplementary material Section 6 enumerates the full derivation in detail. We report that the Laplace approximation with the logistic normal distribution is applied to model the prior, $\boldsymbol{\alpha}$, of the level proportion, $\boldsymbol{\eta}$. We choose a conjugate prior of a multinomial, so $p(\boldsymbol{\eta}_n|\boldsymbol{\alpha})$ follows the Dirichlet distribution. To configure the inference network on the Dirichlet prior, the Laplace approximation is used \citep{mackay1998choice,srivastava2017autoencoding,hennig2012kernel}. 
\begin{align}
& \mathscr{L}(\boldsymbol{x})= \mathbb{E}_{q}\biggl[ \log \frac{p(\boldsymbol{v})}{q(\boldsymbol{v}|\boldsymbol{x})} + \log \frac{p(\boldsymbol{\eta})}{q(\boldsymbol{\eta}|\boldsymbol{x})} \nonumber \\
& + \log \prod_{\boldsymbol{\zeta},\boldsymbol{l}} \frac{p(\boldsymbol{\zeta}|\boldsymbol{v})}{q(\boldsymbol{\zeta}|\boldsymbol{x})} \frac{p(\boldsymbol{l}|\boldsymbol{\eta})}{q(\boldsymbol{l}|\boldsymbol{x})} \frac{p(\boldsymbol{z}|\boldsymbol{\mu}_{\boldsymbol{\zeta}_{\boldsymbol{l}}},\boldsymbol{\sigma}_{\boldsymbol{\zeta}_{\boldsymbol{l}}}^2)}{q(\boldsymbol{z}|\boldsymbol{x})}+\log p(\boldsymbol{x}|\boldsymbol{z}) \biggr] \nonumber \\[7pt]
= & \textstyle{ 
\; \sum_{i\in \mathscr{M}_{T}} [ \log \gamma + ( \gamma -1) (\psi (b_i) - \psi (a_i+b_i)) -  } \nonumber \\
&  \textstyle{ 
  \{ \log \Gamma(a_i+b_i) - \log \Gamma(a_i) - \log \Gamma(b_i) + (a_i-1)\psi(a_i)  }  \nonumber \\
& \textstyle{ 
 + (b_i-1)\psi(b_i) \} ] + \sum_{n=1}^{N} [ \mathbb{E}_{q}[\log p(\boldsymbol{\zeta}_n|\boldsymbol{v})]   }  \nonumber \\
& \textstyle{ 
 + \sum_{i=1}^L(\alpha_i-1)(\psi(\widetilde{\alpha}_{ni})-\psi(\widetilde{\alpha}_{n0})) + \log \Gamma(\alpha_0)  } \nonumber \\
& \textstyle{
- \sum_{i=1}^{L}\log \Gamma(\alpha_i) + \sum_{l'}\omega_{nl'}  (\psi(\widetilde{\alpha}_{nl^{'}})-\psi(\widetilde{\alpha}_{n0}))  } \nonumber \\
& \textstyle{ 
+ \sum_{\boldsymbol{\zeta'}}S_{n\boldsymbol{\zeta'}}  \biggl\{ \sum_{l'}\omega_{nl'}  \biggl(\sum_{j=1}^J -\frac{1}{2} \log (2\pi\sigma^2_{\zeta'_{nl'},j}) } \nonumber \\
& \textstyle{ 
- \frac{(\widetilde{\mu}_{z_{nj}} - \mu_{\zeta'_{nl'},j})^2}{2\sigma^2_{\zeta'_{nl'},j}}
- \frac{\widetilde{\sigma}_{z_{nj}}^2}{2\sigma^2_{\zeta'_{nl'},j}}  \biggr)   \biggr\} 
} \nonumber \\
& \textstyle{ 
+ \frac{1}{R}\sum_{r=1}^{R}\sum_{d=1}^{D} -\frac{1}{2} \log (2\pi \sigma_{x_{nd}}^{{(r)}^{2}}) - \frac{(x_{nd}-\mu_{x_{nd}}^{(r)})^2}{2\sigma_{x_{nd}}^{(r)^{2}}}  } \nonumber \\
& \textstyle{
 - \{ \sum_{\boldsymbol{\zeta'}} \frac{S_{n\boldsymbol{\zeta'}}}{\sum_{\boldsymbol{\zeta''}}S_{n\boldsymbol{\zeta''}}}\log \frac{S_{n\boldsymbol{\zeta'}}}{\sum_{\boldsymbol{\zeta''}}S_{n\boldsymbol{\zeta''}}} + \log \Gamma(\widetilde{\alpha}_{n0}) } \nonumber \\
& \textstyle{
 - \sum_{i=1}^L \log \Gamma(\widetilde{\alpha}_{ni}) + \sum_{i=1}^L (\widetilde{\alpha}_{ni}-1)(\psi(\widetilde{\alpha}_{ni})-\psi(\widetilde{\alpha}_{n0}))  } \nonumber \\
& \textstyle{
+ \sum_{l'}\omega_{nl'}  \log \omega_{nl'} } \nonumber \\
& \textstyle{
-\frac{J}{2}\log(2\pi) -\frac{1}{2}\sum_{j=1}^{J}(1+\log \widetilde{\sigma}^2_{z_{nj}}) \} ]}
\label{eq:hcrl}
\end{align}
where $\widetilde{\alpha}_{n0} = \sum_{i=1}^{L}\widetilde{\alpha}_{ni}$, $\alpha_0 = \sum_{i=1}^{L}\alpha_{i}$, $\psi$ denotes the digamma function, and $R$ is the mini-batch size.

\newpage
\subsection{Training Algorithm of Clustering Hierarchy}
HCRL is formalized according to the stick-breaking process scheme. Unlike the CRP, the stick-breaking process does not represent the direct sampling of the mixture component at the data instance level. Therefore, it is necessary to devise a heuristic algorithm for operations, such as \textit{GROW}, \textit{PRUNE}, and \textit{MERGE}, to refine the hierarchy structure. supplementary material Section 3 provides details about each operation. In the below description, an \textit{\text{inner}} path and a \textit{\text{full}} path refer to the path ending with an internal node and a leaf node, respectively.

\begin{algorithm}[htbp]
	\caption{Training for Hierarchically Clustered Representation Learning}	
	\label{alg:hcrl}	
	\begin{algorithmic}[1]		
		\REQUIRE Training data $\boldsymbol{x}$; number of epochs, $E$; tree-based hierarchy depth, $L$; period of performing GROW, $t_{\text{grow}}$; minimum number of epochs locking the hierarchy, $t_{\text{lock}}$
		\ENSURE ${{T}}^{(E)}, \boldsymbol{\omega}, \{a_{i}, b_{i}, \boldsymbol{\mu}_{i}, \boldsymbol{\sigma}_{i}^{2}\}_{i\in \mathscr{M}_{{T}^{(E)}}}$
		
		%
		%
		
		\STATE $\boldsymbol{\mu}_{\overline{\boldsymbol{\zeta}}_{1:L}}, \boldsymbol{\sigma}_{\overline{\boldsymbol{\zeta}}_{1:L}}^{2} \leftarrow$ Initialize $L$ Gaussian of a single path $\overline{\boldsymbol{\zeta}}$
		\STATE ${T}^{(0)} \leftarrow $ Initialize the tree-based hierarchy having $\overline{\boldsymbol{\zeta}}$
		\STATE $t \leftarrow 0 $ 
		\FOR {each epoch $e=1, \cdots,E$}
	    \STATE Update the weight parameters using $\nabla \mathscr{L}(\boldsymbol{x})$
		\STATE $\{a_{i}, b_{i}, \boldsymbol{\mu}_{i}, \boldsymbol{\sigma}_{i}^{2}\}_{i\in \mathscr{M}_{{T}^{(e-1)}}} \leftarrow$ Update node-specific parameters using $\nabla_{\boldsymbol{a},\boldsymbol{b},\boldsymbol{\mu},\boldsymbol{\sigma}^2} \mathscr{L}(\boldsymbol{x})$ 
		\STATE Update other variational parameters using $\nabla\mathscr{L}(\boldsymbol{x})$ \label{alg:update2}
		\IF {$\textup{mod}(e,t_{\text{grow}}) = 0$} 
		\STATE ${{T}}^{(e)},\sQ$ $\leftarrow$ GROW
		\ENDIF
		\IF {${{T}}^{(e)} = {{T}}^{(e-1)}$ and $t \geq t_{\text{lock}}$} 
		\STATE ${{T}}^{(e)},\sQ$ $\leftarrow$ PRUNE
		\IFTHEN{${{T}}^{(e)} = {T}^{(e-1)}$} 
		{${{T}}^{(e)},\sQ$ $\leftarrow$ MERGE} 
		\ENDIF
		\IFTHENELSE {${{T}}^{(e)} \neq {{T}}^{(e-1)}$}
		{$t \leftarrow 0 $}
		{$t \leftarrow t + 1 $}
		\ENDFOR		
	\end{algorithmic}
\end{algorithm}
\begin{itemize}[noitemsep,topsep=0pt]
\item
\textbf{GROW} expands the hierarchy by creating a new branch under the heavily weighted internal node. Compared with the work of \citet{wang2009variational}, we modified GROW to first sample a path, $\overline{\boldsymbol{\zeta}}^{*}$, proportional to $\sum_{n}q(\boldsymbol{\zeta}_{n}=\overline{\boldsymbol{\zeta}}^{*})$, and then to grow the path if the sampled path is an inner path.
\item \textbf{PRUNE} cuts a randomly sampled minor full path, $\overline{\boldsymbol{\zeta}}^{*}$, satisfying $\frac{\sum_{n}q(\boldsymbol{\zeta}_{n}=\overline{\boldsymbol{\zeta}}^{*})}
{\sum_{n,\overline{\boldsymbol{\zeta}}}q(\boldsymbol{\zeta}_{n}=\overline{\boldsymbol{\zeta}})} < \delta$, where $\delta$ is the pre-defined threshold. If the removed leaf node of the full path is the last child of the parent node, we also recursively remove the parent node.
\item \textbf{MERGE} combines two full paths, $\overline{\boldsymbol{\zeta}}^{(i)}$ and $\overline{\boldsymbol{\zeta}}^{(j)}$, with similar posterior probabilities, measured by $ J(\overline{\boldsymbol{\zeta}}^{(i)},\overline{\boldsymbol{\zeta}}^{(j)}) = \boldsymbol{q}_{i}\boldsymbol{q}_{j}^{T}/|\boldsymbol{q}_i||\boldsymbol{q}_j|$, where $\boldsymbol{q}_i=[q(\boldsymbol{\zeta}_1=\overline{\boldsymbol{\zeta}}^{(i)}), \cdots ,q(\boldsymbol{\zeta}_N=\overline{\boldsymbol{\zeta}}^{(i)})]$. 
\vspace{-0.2em}
\end{itemize}


Algorithm \ref{alg:hcrl} summarizes the overall algorithm for HCRL. The tree-based hierarhcy $T$ is defined as $(\mathbb{N}, \mathbb{P})$, where $\mathbb{N}$ and $\mathbb{P}$ denote a set of nodes and paths, respectively. We refer to the node at level $l$ lying on path $\boldsymbol{\zeta}$, as $\textup{N} (\boldsymbol{\zeta}_{1:l})\in \mathbb{N}$. The defined paths, $\mathbb{P}$, consist of full paths, $\mathbb{P}_{\text{full}}$, and inner paths, $\mathbb{P}_{\text{inner}}$, as a union set. The GROW algorithm is executed for every specific iteration period, $t_{\text{grow}}$. After ellapsing $t_{\text{lock}}$ iterations since performing the GROW operation, we begin to check whether the PRUNE or MERGE operation should be performed. We prioritize the PRUNE operation first, and if the condition of performing PRUNE is not satisfied, we check for the MERGE operation next. After performing any operation, we initialize $t$ to 0, which is for locking the changed hierarchy during minimum $t_{\text{lock}}$ iterations to be fitted to the training data.

\section{Experiments}

\begin{table*}[htbp]
	\caption{Test set performance of the negative log likelihood (NLL) and the reconstruction errors (REs).
		Replicated ten times, and the best in bold. $P^{\dagger} < 0.05$ (Student's t-test). \textit{Model-}$L\#$ means that the model trained with the $\#$-depth hierarchy.}
	\label{tbl:ae}
	\begin{center}
		{
			\begin{tabular}{lrrrrrrrr}
				\hline 
				& \multicolumn{2}{c}{\bf MNIST} & \multicolumn{2}{c}{\bf CIFAR-100} & \multicolumn{2}{c}{\bf RCV1\_v2} & \multicolumn{2}{c}{\bf 20Newsgroups} \\
				\cline{2-9}
				\textbf{Model} & \multicolumn{1}{c}{\bf NLL} & \multicolumn{1}{c}{\bf REs} & \multicolumn{1}{c}{\bf NLL} & \multicolumn{1}{c}{\bf REs} & \multicolumn{1}{c}{\bf NLL} & \multicolumn{1}{c}{\bf REs} & \multicolumn{1}{c}{\bf NLL} & \multicolumn{1}{c}{\bf REs} \\ \hline
				
				VAE & $230.71$ & $10.46$ & $1960.06$ & $57.54$ 
				& $2559.46$ & $1434.59$ & $2735.80$ & $1788.22$\\
				VaDE & $217.20$ & $10.35$ & $1921.85$ & $53.60$ 
				& $2558.32$ & $1426.38$ & $2733.46$ & $1782.86$\\
				IDEC & N$\mathbin{/}$A & $12.75$ & N$\mathbin{/}$A & $64.09$ 
				& N$\mathbin{/}$A & $1376.26$ & N$\mathbin{/}$A & $\boldsymbol{1660.61}^{\dagger}$\\
				DCN & N$\mathbin{/}$A & $11.30$ & N$\mathbin{/}$A & $44.26$ 
				& N$\mathbin{/}$A & $1361.98$ & N$\mathbin{/}$A & $1691.17$\\
				IMVAE & $296.57$ & $10.69$ & $1992.83$ & $\boldsymbol{40.45}^{\dagger}$ 
				& $2566.01$ & $1387.02$ & $2722.81$ & $1718.08$\\ 
				VAE-nCRP-$L3$ & $718.78$ & $32.67$ & $2969.62$ & $198.66$
				& $2642.88$ & $1538.42$ & $2712.28$ & $1680.56$ \\ 
				VAE-nCRP-$L4$ & $721.00$ & $32.53$ & $2950.73$ & $198.97$
				& $2646.48$ & $1542.81$ & $2713.58$ & $1680.71$ \\ 
				\hline
				HCRL1-$L3$ & $209.59^{\dagger}$ & $9.28^{\dagger}$ & $1864.69^{\dagger}$ & 55.12 & $2562.79$ & $1418.30$ & $2732.10$ & $1792.13$ \\
				HCRL1-$L4$ & $212.31^{\dagger}$ & $8.31^{\dagger}$ & $1860.22 ^{\dagger}$ & 55.56 & $2555.84$ & $1404.23$ & $2727.49$ & $1754.94$  \\
				HCRL2-$L3$ & $\mathbf{203.24}^{\dagger}$ & $8.70^{\dagger}$ & $1843.40^{\dagger}$ & $50.44$
				& $2554.50^{\dagger}$ & $1395.05$ & $2726.75$ & $1828.71$\\
				HCRL2-$L4$ & $203.91^{\dagger}$ & $\mathbf{8.16}^{\dagger}$ 
				& $\mathbf{1849.13}^{\dagger}$ & $50.47$ & $\mathbf{2535.43^{\dagger}}$ & $\mathbf{1353.34}$ & $\mathbf{2702.88}$ & $1711.30$\\
				\hline 
			\end{tabular}
		}
	\end{center}\
	\vspace{-0.5em}
\end{table*}

\subsection{Datasets and Baselines}
\textbf{Datasets:} We used various hierarchically organized benchmark datasets as well as MNIST.
\begin{itemize}[noitemsep,topsep=0pt]
\item
\textbf{MNIST \citep{lecun1998gradient}:} 28x28x1 handwritten image data, with 60,000 train images and 10,000 test images. We reshaped the data to 784-d in one dimension. 
\item
\textbf{CIFAR-100 \citep{krizhevsky2009learning}:} 32x32x3 colored images with 20 coarse and 100 fine classes. We used 3,072-d flattened data with 50,000 training and 10,000 testing. 
\item
\textbf{RCV1\_v2 \citep{lewis2004rcv1}:} The preprocessed text of the Reuters Corpus Volume. We preprocessed the text by selecting the top 2,000 tf-idf words. We used the hierarchical labels up to the 4-level, and the multi-labeled documents were removed. The final preprocessed corpus consists of 11,370 training and 10,000 testing documents randomly sampled from the original test corpus.  
\item
\textbf{20Newsgroups \citep{lang1995newsweeder}:} The benchmark text data extracted from 20 newsgroups, consisting 11,314 training and 7,532 testing documents. We also labeled by 4-level following the annotated hierarchical structure. We preprocessed the data through the same process as that of RCV1\_v2. 
\end{itemize}



\textbf{Baselines:} 
We completed our evaluation in two aspects: 1) optimizing the density estimation, and 2) clustering the hierarchical categories. First, we evaluated HCRL1 and HCRL2 from the density estimation perspective by comparing it with diverse flat clustered representation learning models, and VAE-nCRP. Second, we tested HCRL1 and HCRL2 from the accuracy perspective by comparing it with multiple divisive hierarchical clusterings. The below is the list of baselines. We also added the two-stage pipeline approaches, where we trained features from VaDE first and then applied the hierarchical clusterings. We reused the open source codes\footnote{https://github.com/XifengGuo/IDEC (IDEC); \\ https://github.com/boyangumn/DCN (DCN); \\ https://github.com/prasoongoyal/bnp-vae (VAE-nCRP); \\ http://vision.jhu.edu/code/ (SSC-OMP)} provided by the authors for several baselines, such as IDEC, DCN, VAE-nCRP, and SSC-OMP.
\begin{enumerate}[noitemsep,topsep=0pt]
\item \textbf{Variational Autoencoder (VAE) \citep{kingma2013auto}:} places a single Gaussian prior on embeddings. 
\item \textbf{Variational Deep Embedding (VaDE) \citep{jiang2016variational}:} jointly optimizes a Gaussian mixture model and representation learning. 
\item \textbf{Improved Deep Embedded Clustering (IDEC) \citep{guo2017improved}:} improves DEC \citep{xie2016unsupervised} by attatching decoder structure. We use the code by the authors.
\item \textbf{Deep Clustering Network (DCN) \citep{yang2017towards}:} optimizes the K-means-related cost defined on the embedding space. We used the open source code provided by the authors.
\item \textbf{Infinite Mixture of Variational Autoencoders (IMVAE) \citep{abbasnejad2017infinite}:} searches for the infinite embedding space by using a Bayesian nonparametric prior.
\item \textbf{Variational Autoencoder - nested Chinese Restaurant Process (VAE-nCRP) \citep{goyal2017nonparametric}:} We used the open source code provided by the authors.
\item \textbf{Hierarchical K-means (HKM) \citep{nister2006scalable}:} performs K-means \citep{lloyd1982least} recursive in a top-down way.
\item \textbf{Mixture of Hierarchical Gaussians (MOHG) \citep{vasconcelos1999learning}:} infers the level-specific mixture of Gaussians.
\item \textbf{Recursive Gaussian Mixture Model (RGMM):} runs GMM recursively in a top-down manner.
\item \textbf{Recursive Scalable Sparse Subspace Clustering by Orthogonal Matching Pursuit (RSSCOMP):} performs SSCOMP \citep{you2016scalable} recursively for hierarchical clustering. SSCOMP is a well-known methods for image clustering, and we used the open source code.
\end{enumerate}

\renewcommand{\tabcolsep}{4.5pt}
\begin{table}[h]
\small
\caption{Hierarchical clustering accuracies with F-scores, on CIFAR-100 with a depth of three, RCV1\_v2 with a depth of four, and 20Newsgroups with a depth of four. Replicated ten times, and a confidence interval with 95\%. Best in bold.}
\label{tbl:hc}
\begin{center}
\begin{tabular}{lrrr}
\hline 
\textbf{Model} &\textbf{CIFAR-100} & \textbf{RCV1\_v2} & \textbf{20Newsgroups} \\ \hline
HKM & $0.162_{\pm 0.008}$ & $0.256_{\pm 0.068}$ & $0.410_{\pm 0.043}$ \\
MOHG & $0.085_{\pm 0.038}$ & $0.103_{\pm 0.014}$ & $0.040_{\pm 0.012}$ \\
RGMM & $0.169_{\pm 0.012}$ & $0.274_{\pm 0.052}$ & $0.435_{\pm 0.037}$ \\
RSSCOMP & $0.146_{\pm 0.023}$ & $0.266_{\pm 0.055} $ & $0.295_{\pm 0.047}$ \\
VAE-nCRP & $0.201_{\pm 0.008}$ & $0.413_{\pm 0.024} $ & $0.558_{\pm 0.027}$ \\
\hline
VaDE+HKM &  $0.164_{\pm 0.012}$ & $0.331_{\pm 0.066}$ & $0.485_{\pm 0.056}$ \\
VaDE+MOHG & $0.166_{\pm 0.016}$ & $0.423_{\pm 0.093}$ & $0.492_{\pm 0.071}$ \\
VaDE+RGMM &  $0.181_{\pm 0.013}$ & $0.386_{\pm 0.062}$ & $0.410_{\pm 0.065}$ \\
\scriptsize VaDE+RSSCOMP & $0.192_{\pm 0.021}$ & $0.272_{\pm 0.044}$ & $0.291_{\pm 0.043}$ \\
\hline
HCRL1 & $0.199_{\pm 0.016}$  & $0.437_{\pm 0.029}$ & $0.566_{\pm 0.048}$  \\
HCRL2 & $\mathbf{0.225}_{\pm 0.014}$ & $\mathbf{0.455}_{\pm 0.030}$ & $\mathbf{0.601}_{\pm 0.097}$ \\
\hline 
\end{tabular}
\end{center}
\vspace{-1.5em}
\end{table}

\subsection{Quantitative Analysis}

We used two measures to evaluate the learned representations in terms of the density estimations: 1) negative log likelihood (NLL), and 2) reconstruction errors (REs). Autoencoder models, such as IDEC and DCN, were tested only for the REs. The NLL is estimated with 100 samples. Table \ref{tbl:ae} indicates that HCRL is best in the NLL and is competent in the REs which means that the hierarchically clustered embeddings preserve the intrinsic raw data structure.

VaDE generally performed better than VAE did, whereas other flat clustered representation learning models tended to be slightly different for each dataset. HCRL1 and HCRL2 showed better results with a deeper hierarchy of level four than of level three, which implies that capturing the deeper hierarchical structure is likely to be useful for the density estimation, and especially HCRL2 showed overall competent performance.

Additionally, we evaluated hierarchical clustering accuracies by following \citet{xie2016unsupervised}, except for MNIST that is flat structured. 
Table \ref{tbl:hc} points out that HCRL2 has better micro-averaged F-scores compared with every baseline. HCRL2 is able to reproduce the ground truth hierarchical structure of the data, and this trend is consistent when HCRL2 compared with the pipelined model, such as VaDE with a clustering model. The result of the comparisons with the clustering models, such as HKM, MOHG, RGMM, and RSSCOMP, is interesting because it experimentally proves that the joint optimization of hierarchical clustering in the embedding space improves hierarchical clustering accuracies. HCRL2 also presented better hierarchical accuracies than VAE-nCRP. We conjecture the reasons for the modeling aspect of VAE-nCRP: 1) the simplified prior modeling on the variance of the mixture component as just constants, and 2) the non-flexible learning of the internal components.

The performance gain of HCRL2 compared to HCRL1 arises from the detailed modeling of the level proportion. The prior assumption that the level proportion is shared by all data may give rise to the optimization biased towards the learning of leaf components. Specifically, a lot of data would be generated from the leaf components with the high probability since the leaf components have small variance, which causes the global level proportion to focus the high probability on the leaf level. This mechanism accelerates the biased optimization to the leaf components, and on the other hand, HCRL2 allows the flexible learning of the level proportions.


\subsection{Qualitative Analysis}

\begin{figure*}[htbp]
\centerline{\includegraphics[width=0.99\textwidth]{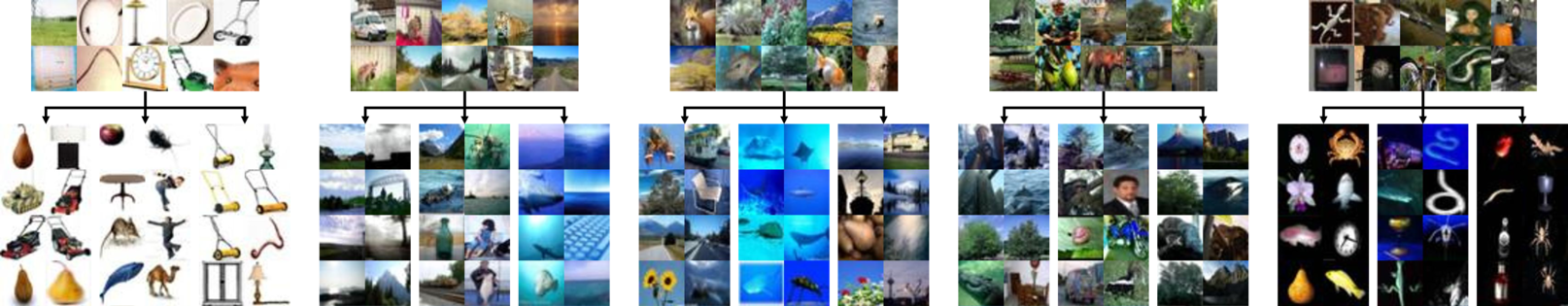}}
\caption{Example extracted sub-hierarchies on CIFAR-100}
\label{fig:cifar}
\end{figure*}

\begin{figure*}[htbp]
    \centering
    \begin{subfigure}{0.175\textwidth}
	\centering
	\captionsetup{justification=centering}
        \includegraphics[width=\textwidth]{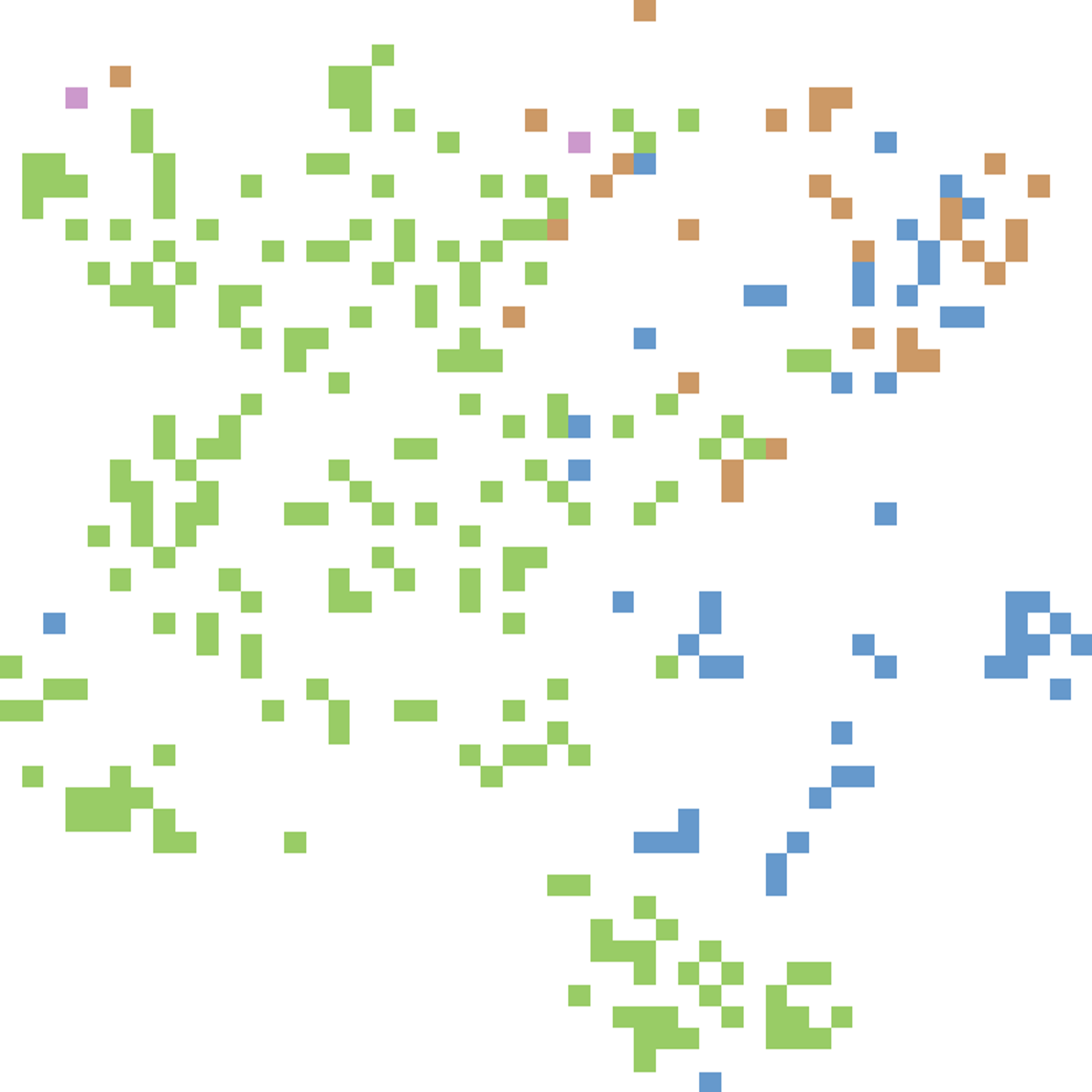}
        \caption{VAE \citep{kingma2013auto}}
    \end{subfigure}
    \quad
    \begin{subfigure}{0.175\textwidth}
	\centering
	\captionsetup{justification=centering}
        \includegraphics[width=\textwidth]{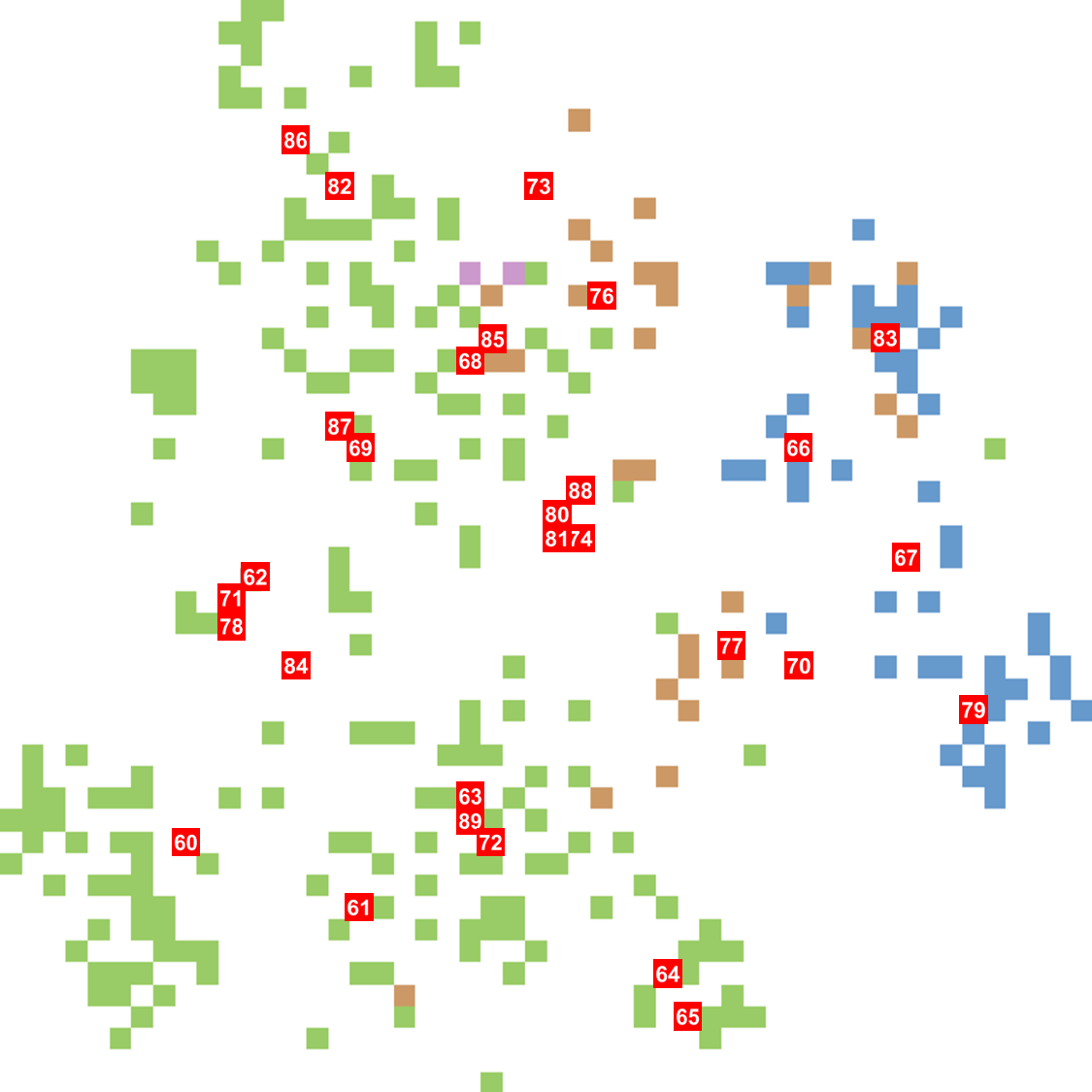}
        \caption{VaDE \newline \citep{jiang2016variational}}
    \end{subfigure}
    \quad
    \begin{subfigure}{0.175\textwidth}
	\centering
	\captionsetup{justification=centering}
        \includegraphics[width=\textwidth]{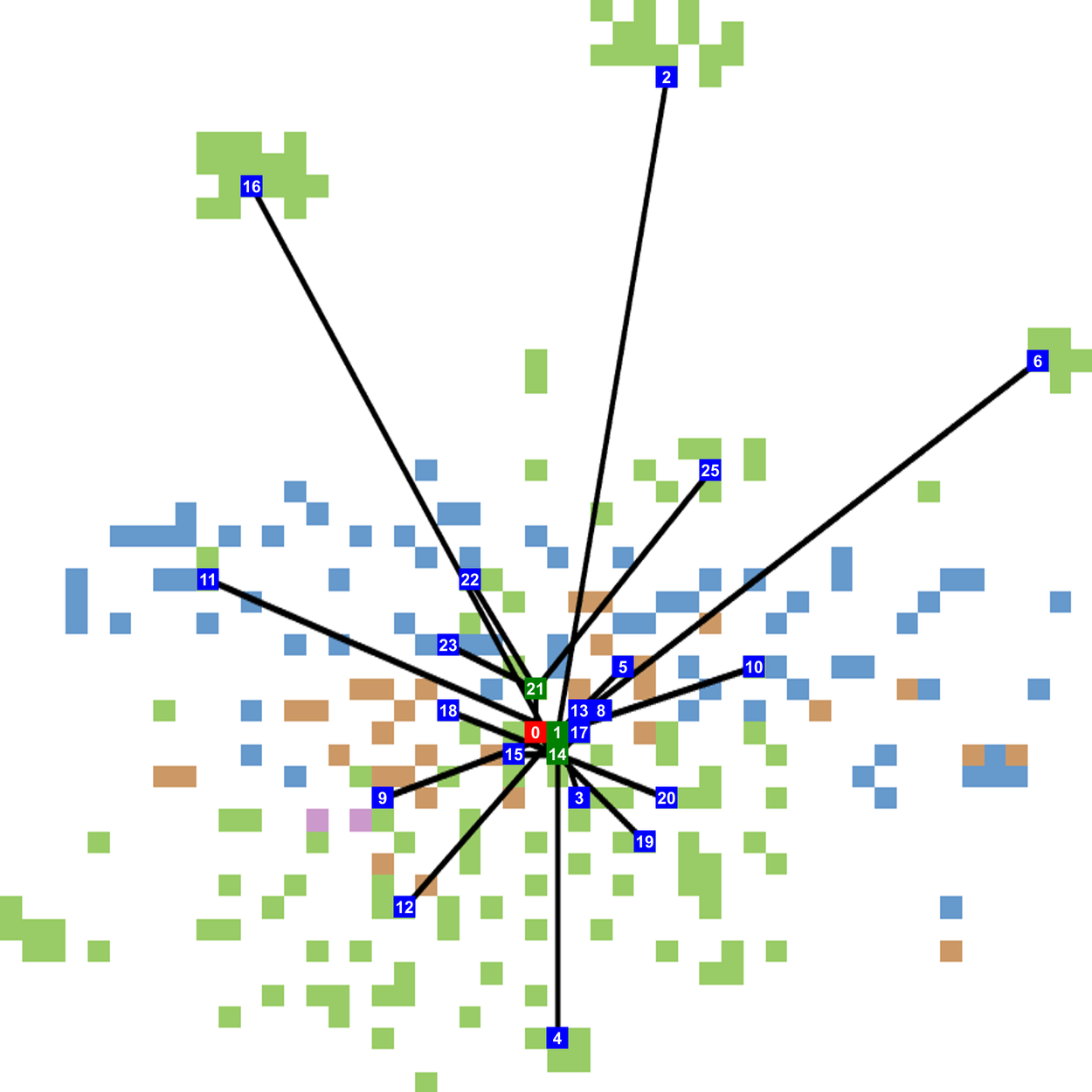}
        \caption{VAE-nCRP \newline \citep{goyal2017nonparametric}}
    \end{subfigure}
    \quad
    \begin{subfigure}{0.175\textwidth}
        \includegraphics[width=\textwidth]{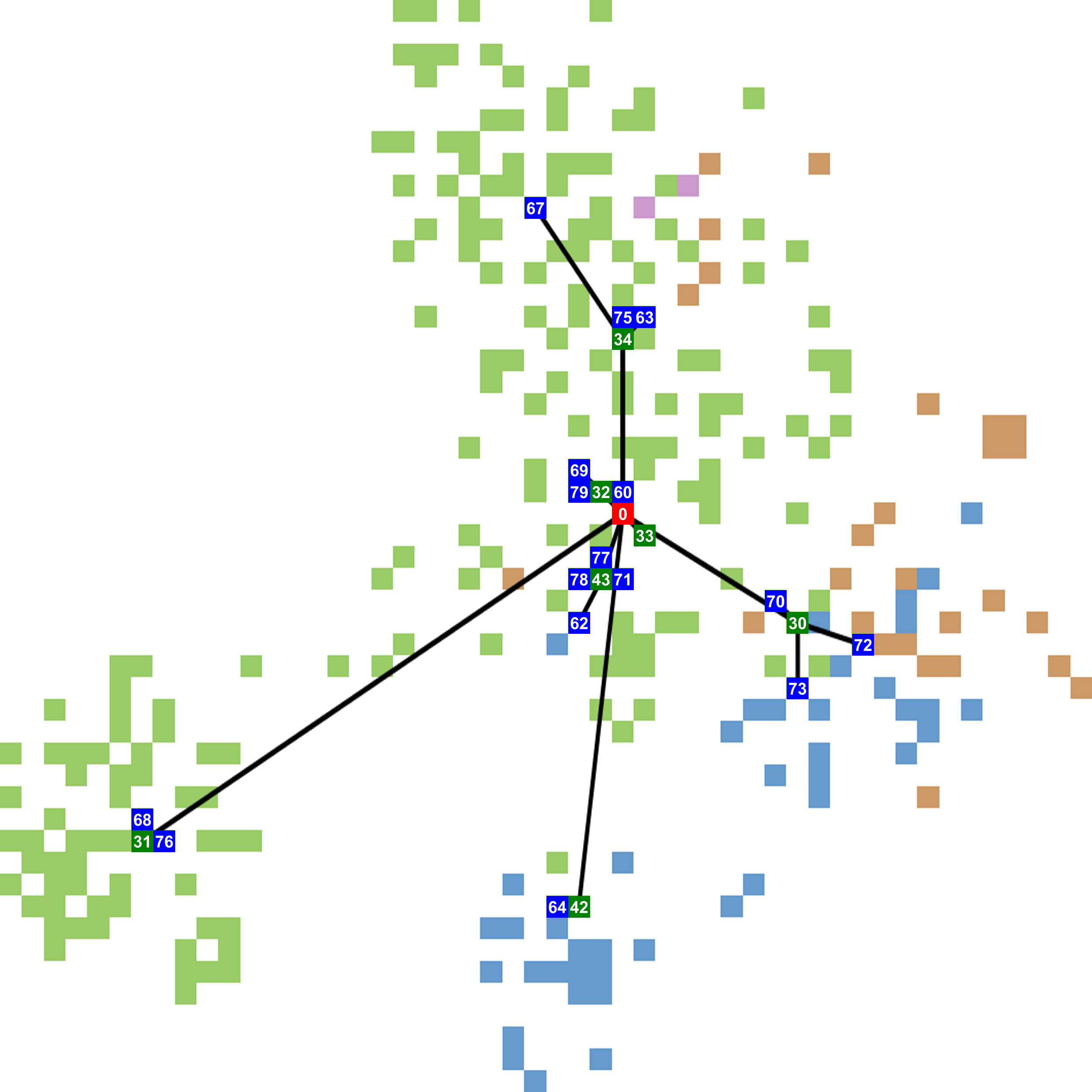}
        \caption{HCRL1}
    \end{subfigure}
    \quad
    \begin{subfigure}{0.175\textwidth}
        \includegraphics[width=\textwidth]{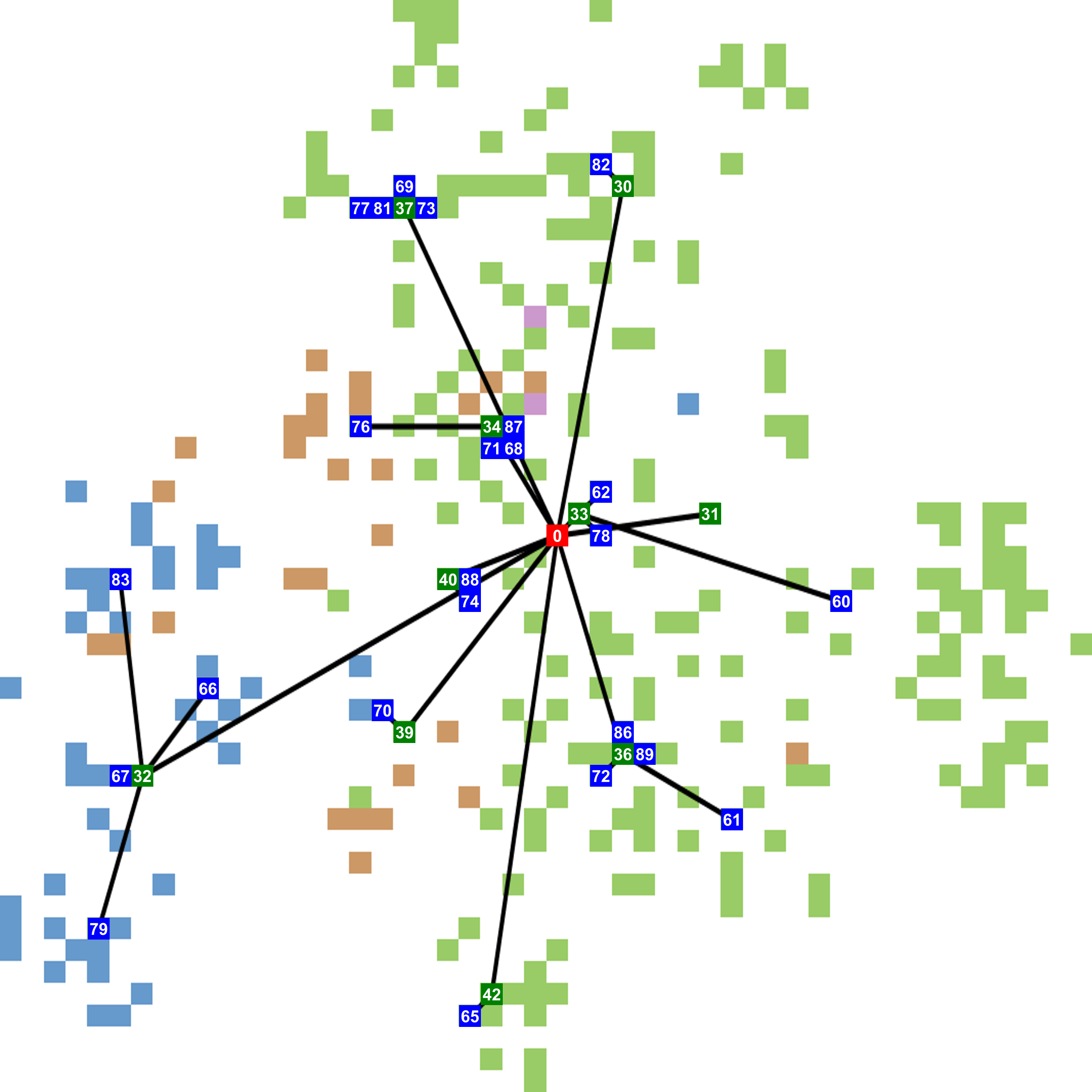}
        \caption{HCRL2}
    \end{subfigure}
    \caption{Comparison of embeddings on RCV1\_v2, plotted using t-SNE \citep{maaten2008visualizing}. We mark the mean of a mixture component with a numbered square, colored in \{red\} for VaDE, \{red (root), green (internal), blue (leaf)\} for VAE-nCRP, HCRL1, and HCRL2. 
The first-level sub-hierarchies are indicated with four colors.}
    \label{fig:rcv}
\end{figure*}


\begin{figure}[htbp]
\centerline{\includegraphics[width=0.55\columnwidth]{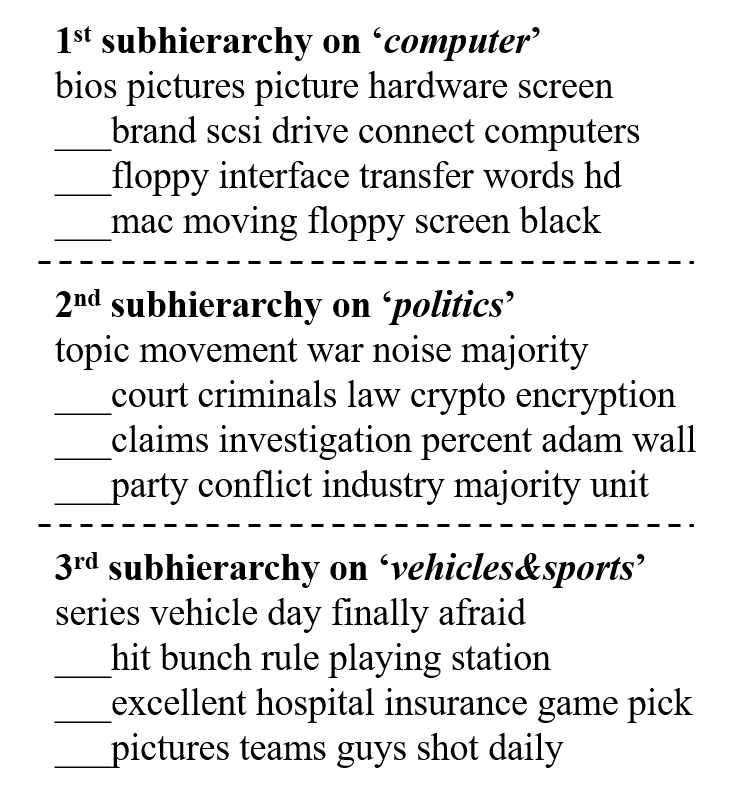}}
     \caption{Example extracted sub-hierarchies on 20Newsgroups}
    \label{fig:20news}
\end{figure}

\textbf{MNIST: } 
In Figure \ref{fig:mnist}, the digits \{4, 7, 9\} and the digits \{3, 8\} are grouped together with a clear hierarchy, which was consistent between HCRL2 and VaDE. Also, some digits \{0, 4, 2\} in a round form are grouped, together, in HCRL2. In addition, among the reconstructed digits from the hierarchical mixture components, the digits generated from the root have blended shapes from 0 to 9, which is natural considering the root position.

\textbf{CIFAR-100: } 
Figure \ref{fig:cifar} shows the hierarchical clustering results on CIFAR-100, which are inferred from HCRL2. Given that there were no semantic inputs from the data, the color was dominantly reflected in the clustering criteria. However, if one observes the second hierarchy, the scene images of the same sub-hierarchy are semantically consistent, although the background colors are slightly different. 

\textbf{RCV1\_v2: } 
Figure \ref{fig:rcv} shows the embedding of RCV1\_v2. VAE and VaDE show no hierarchy, and close sub-hierarchies are distantly embedded. Since the flat clustered representation learning focuses on isolating clusters from each other, the distances between different clusters tend to be uniformly distributed. VAE-nCRP guides the internal mixture components to be agglomerated at the center, and the cause of agglomeration is the generative process of VAE-nCRP, where the parameter of the internal components are inferred without direct information from data. 

HCRL1 and HCRL2 show a relatively clear separation between the sub-hierarchy without the agglomeration. However, HCRL2 is significantly superior to HCRL1 in terms of learning the hierarchically clustered embeddings. In HCRL1, the distant embeddings are learned even though they belong to the same sub-hierarchy. 

\textbf{20Newsgroups: } 
Figure \ref{fig:20news} shows the example sub-hierarchies on 20Newsgroups. We enumerated topic words from documents with top-five likelihoods for each cluster, and we filtered the words by tf-idf values. We observe relatively more general contents in the internal clusters than in the leaf clusters of each internal cluster.

\section{Conclusion}
In this paper, we have presented a hierarchically clustered representation learning framework for the hierarchical mixture density estimation on deep embeddings. HCRL aims at encoding the relations among clusters as well as among instances to preserve the internal hierarchical structure of data. We have introduced two models called HCRL1 and HCRL2, whose the main differentiated features are 1) the crucial assumption regarding the internal mixture components for having the ability to generate data directly, and 2) the level selection modeling. HCRL2 improves the performance of HCRL1 by inferring the data-specific level proportion through the unbalanced autoencoding neural architecture. From the modeling and the evaluation, we found that our proposed models enable the improvements due to the high flexibility modeling compared with the baselines.


\newpage

\section*{Acknowledgements}


\nocite{langley00}

\bibliographystyle{icml2019}
\bibliography{icml2019_hcrl}

%
%
%

\end{document}